\documentclass{article}
\usepackage{spconf,amsmath,graphicx,hyperref}

\usepackage{amssymb} 
\usepackage{pifont} 

\usepackage{array} 
\usepackage{color, xcolor} 
\usepackage{colortbl} 
\usepackage{booktabs} 
\usepackage{multicol} 
\usepackage{multirow} 
\usepackage{adjustbox} 
\usepackage{subcaption} 

\definecolor{LightBlue}{rgb}{0.94,0.97,1}
\newcommand{\modelname}{CERF}

\title{CERF: Communication-Efficient and Retraining-Free Collaborative Perception}
\name{Jiuwu Hao$^{1,2,*}$, 
Ziyi Ni$^{1,2,*}$, 
Liguo Sun$^{2}$, 
Yuting Wan$^{1,2}$, 
Yueyang Wu$^{1 ,2}$,
Ti Xiang$^{1,2}$, 
Haolin Song$^{1,2}$, 
Pin Lv$^{2,\dagger}$ 
\thanks{$^{*}$~Equal contribution. \quad $^{\dagger}$~Corresponding author.}}

\address{$^{1}$School of Artificial Intelligence, University of Chinese Academy of Sciences \\
$^{2}$Institute of Automation, Chinese Academy of Sciences}

\begin{document}
\maketitle
\begin{abstract}
Collaborative perception shares information among multiple agents to obtain a comprehensive scene representation, enhancing the perceptual capability of individual agents.
However, most existing methods rely on transmitting and fusing dense feature maps for collaboration, which incurs inevitable communication overhead and heterogeneity challenges, limiting their practicality for real-world deployment.
To address these challenges, we propose \textbf{CERF}, a novel \textbf{C}ommunication-\textbf{E}fficient and \textbf{R}etraining-\textbf{F}ree framework for open heterogeneous collaborative perception.
In {\modelname}, we introduce a new virtual modality (termed Poture), which is generated from the perception outputs of other agents, to augment the extracted Bird's Eye View (BEV) features of the ego agent.
To mitigate transmission delays, we employ a Kalman-filter based tracker and a motion forecasting model to derive the current predictions from historical perception results.
Extensive experiments demonstrate that {\modelname} achieves performance comparable to mainstream intermediate-collaboration methods while reducing communication overhead by $95\%$ across various downstream tasks.
Furthermore, {\modelname} enables seamless integration of unknown heterogeneous agents into the existing collaborative framework without additional retraining costs.
Code is available at \href{https://github.com/uestchjw/CERF}{\textcolor{purple}{https://github.com/uestchjw/CERF}}.
\par
\end{abstract}

\section{Introduction}
Collaborative perception has emerged as a promising paradigm for enhancing situational awareness of individual agents in complex scenarios \cite{DHD}. It extends the perception range and mitigate the occlusion issue by exchanging perceptual information among connected agents.
Recent studies have demonstrated that collaborative perception can be applied to various computer vision tasks based on Unmanned Aerial Vehicles (UAVs), including 3D object detection \cite{Griffin,AGC-Drive}, object tracking \cite{UAV3D,U2UData}, and trajectory prediction \cite{DHD}.
\par
Communication strategy is a crucial concern in collaborative perception.
Despite tremendous progress in model design and perception performance, most previous work focus on intermediate-collaboration strategy, which transmits and fuses implicit feature maps, posing significant challenges in real-world deployments.
Firstly, intermediate features are inherently high-dimensional representations, which can lead to substantial memory costs and transmission overhead.
Extensive efforts have been made to address this issue, including leveraging compressed features \cite{OPV2V,Where2comm,QuantV2X} or sparse queries \cite{TransIFF, yang2025discretization, IFTR}.
However, compared to compact perception outputs, neural features remain a resource-intensive medium.
Secondly, intermediate-collaboration is difficult to seamlessly accommodate unknown agents into the existing collaboration system due to the semantic gap between feature maps extracted by different agents.
In the real world, agents often differ in input modalities, model architectures, and manufacturers, making it infeasible to encompass all agent types during the training stage.
Most prior work use adapters to transform features into a unified space to address the heterogeneity issue \cite{HEAL,Hetecooper,STAMP}.
However, this approach incurs additional retraining costs and raises privacy and security concerns.
\par
To address these challenges, we propose {\modelname}, a novel communication-efficient and retraining-free collaborative perception framework.
Our core insight is that the perception results of other agents already contain sufficient complementary information for the ego agent, making it unnecessary to rely on high-dimensional and abstract neural features.
In light of this, we construct a virtual modality, Poture, from the received perception outputs to augment the ego agent’s local BEV features.
To mitigate transmission delays, we employ a Kalman-filter based tracker to associate historical detections and use motion forecasting to generate predictions for the current timestamp.
Finally, a fused feature map of both Poture and BEV features is fed to various downstream tasks. 
\par
\begin{figure*}[htbp]
  \centering
  \includegraphics[width=\linewidth, keepaspectratio]{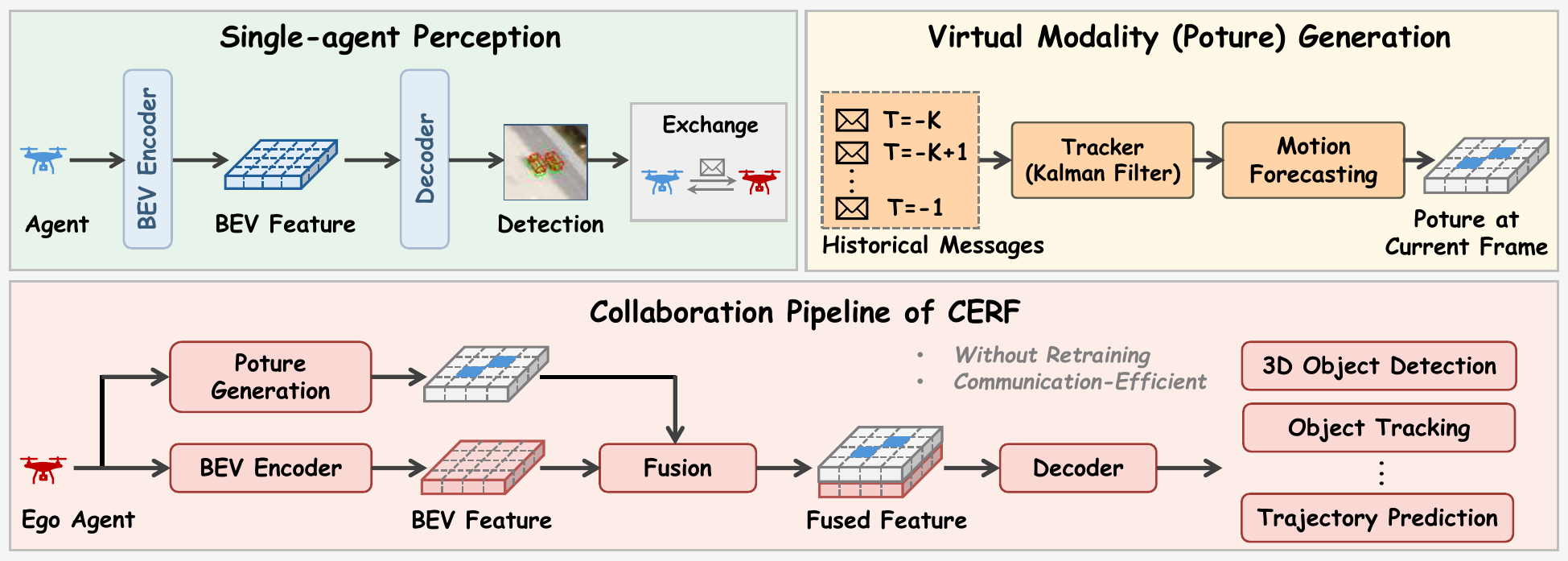}
  \caption{The overview of our proposed {\modelname} framework. In {\modelname}, agents exchange only compact perception results. The ego agent employs a Kalman filter–based tracker and a motion forecasting model to obtain the current prediction from historical messages. Then, the generated virtual modality Poture is fused with the local BEV features and fed to downstream tasks.}
  \label{figure:overview}
\end{figure*}
Extensive experiments on the UAV3D \cite{UAV3D} and Air-Co-Pred \cite{DHD} datasets demonstrate that i) {\modelname} achieves competitive performance with a $95\%$ reduction in communication bandwidth;
ii) CERF can accommodate unknown heterogeneous agents without additional retraining costs.
These results highlight the potential of a practical multi-agent collaborative perception system for real-world deployment.

\section{Problem formulation}
Given $N$ agents in the scene, each is equipped with onboard sensors and capable of both sending and receiving messages.
They may differ in input modalities, model architectures, and downstream tasks.
Notably, among all the $N$ agents, only $N_k$ agents are engaged in the collaborative training phase, with the remaining $N_u$ agents, whose types are unknown, appearing only during the inference stage.
\par
For the $i$th agent, let $O_i$ and $u_i$ represent the raw input and the ground-truth, respectively, and $M_{j \rightarrow i}$ is the message transmitted from agent $j$ to agent $i$.
The goal of a collaborative perception system is to optimize the perceptual performance under communication bandwidth constraints:
\begin{align}
    \mathop {\max }\limits_\theta  \sum\limits_{i = 1}^N &g\left( {{\Phi _\theta }\left( {{O_i},\left\{ {{M_{j \to i}}} \right\}_{j = 1}^{{N_k}},\left\{ {{M_{j \to i}}} \right\}_{j = 1}^{{N_u}}} \right),{u_i}} \right), \\
    \text{s.t.} &\sum\limits_{j = 1}^N {\left| {{M_{j \to i}}} \right|}  \leqslant B
\end{align}
where $g\left(\cdot  \right)$ represents the evaluation metric, ${{\Phi _\theta }}$ is the collaboration model parametrized by $\theta$, and $B$ denotes the communication budget.
\par

\section{METHODOLOGY}

\subsection{Overview}

The overall architecture of the proposed {\modelname} is shown in Fig.~\ref{figure:overview}.
Each agent first takes its local observations as input and outputs the BEV feature maps, and then a decoder is used to obtain the individual perception outputs.
These compact perception results, along with the pose information, will be transmitted to the connected agents.
Upon receiving the message, each agent converts these detections into its own coordinate system through spatial transformation.
To alleviate transmission delays, we use Kalman filtering and motion forecasting to obtain the current predictions from historical frames, generating the virtual modality Poture to enhance the local BEV features.
Finally, the fused feature map is fed to the decoder for diverse downstream tasks.
\par

\subsection{Poture modality generation}
Existing state-of-the-art collaborative perception methods focus on intermediate-fusion pipelines, which involves transmitting and merging dense feature maps.
However, we observe that the compact perception results already contain sufficient complementary information.
Thus, in {\modelname}, the perception outputs serve as the medium to propagate scene information to the ego agent, which can significantly reduce communication overhead and alleviate heterogeneity issues.

\par
Directly merging these detection results, i.e., late-fusion, leads to information loss and degraded performance.
Motivated by MoDAR \cite{MoDAR}, we generate a virtual modality based on the received detections to enhance the raw BEV features.
We name this new modality Poture, \underline{P}erception \underline{o}utputs based fea\underline{ture} map.

\par
For the $i$th agent, the received perception results from agent $j$ at timestamp $t$ consist of a sequence of 3D bounding boxes, which are defined as:
\begin{align}
  P_{ij}^{{t}} &= \left\{ {{b_k}\left| {k \in \left\{ {1,2,...,K} \right\}} \right.} \right\}, \hfill \\
  b &= \left[ {x,y,z,w,h,l,\phi,s} \right] \hfill
\end{align}
where $K$ is the total number of bounding boxes, $\left[ {x,y,z} \right]$ represents the center location, $\left[ {w,h,l} \right]$ is the box size, $\phi$ denotes the heading angle, and the confidence score is expressed by $s$.

\par
After applying the spatial transformation based on the relative pose, we utilize the perception outputs of the previous $K$ frames to predict the detections at the current moment $t_c$, which is:
\begin{equation}
P_{ij}^{{t_c}} = {\Phi _{mf}}\left( {{\Phi _{kf}}\left( {\left\{ {P_{ij}^{{t_{c - n}}}} \right\}_{n = 1}^K} \right)} \right)
\end{equation}
where $\Phi_{kf}$ represents a Kalman-filter based tracker \cite{KF_tracker} used to associate detection results from historical frames, and $\Phi_{mf}$ is a motion forecasting model based on constant velocity.
After obtaining the predicted outputs at the current timestamp, we derive the virtual modality Poture based on the attributes of the bounding boxes:
\begin{equation}
Poture_{ij}^{{t_c}}\left( {x,y} \right) = \left\{ \begin{gathered}
  \left[ {w,h,l,\varphi ,s} \right],{\text{    if }}\left( {x,y} \right){\text{ in }}P_{ij}^{{t_c}} \hfill \\
  0,{\text{                 else}} \hfill \\ 
\end{gathered}  \right.
\end{equation}
After BEV grid discretization (of size $H \times W$), the final virtual modality $Poture \in {\mathbb{R}^{H \times W \times 5}}$ is used to fuse with the local BEV features, delivering complementary information to the ego agent.
\subsection{Poture-BEV fusion}
To accommodate an arbitrary number of collaborative agents, we first apply confidence-aware Non-Maximum Suppression (NMS) to the predicted outputs:
\begin{equation}
Poture_i^{^{{t_c}}} = {\text{NMS}}\left( {Poture_{ij}^{{t_c}}\left| {j \in \left\{ {1,2,...,{N_{{t_c}}}} \right\}} \right.} \right)
\end{equation}
where $N_{t_{c}}$ is the number of connected agents at timestamp $t_c$. At the same location, we retain only the features with high confidence score. Then we concatenate the filtered virtual modality Poture with the original BEV feature $H_i^{{t_c}}$ along the channel dimension, which is:
\begin{equation}
\widehat H_i^{{t_c}} = {\text{Concat}}\left( {Poture_i^{{t_c}},H_i^{{t_c}}} \right)
\end{equation}
where $\widehat H_i^{{t_c}}$ is the fused feature map.
\par
\section{EXPERIMENTS}
\begin{table}[htbp]
\caption{3D object detection results on the UAV3D dataset.}
\label{table: overall_performace}

\centering
\begin{adjustbox}{width=\columnwidth}
\begin{tabular}{c|ccc}
\toprule
\textbf{Method} & \textbf{Bandwidth (Mbps) $\downarrow$}  & \textbf{mAP $\uparrow$} & \textbf{NDS $\uparrow$} \\
\hline
\addlinespace[2.5pt]
No Collaboration & 0.00 & 0.544 & 0.481  \\ 
Late Collaboration & 0.04 & 0.610 & 0.535 \\ 
Early Collaboration & 164.80 & \textbf{0.720} & \textbf{0.608} \\ 
\midrule
Who2com \cite{Who2com} & 19.02 & 0.546 & 0.440 \\
When2com \cite{When2com} & 32.00 & 0.550 & 0.442 \\
V2VNet \cite{V2Vnet} & 11.31 & 0.647 & 0.508 \\ 
Where2comm \cite{Where2comm} & 19.53 & 0.660 & 0.571 \\
DiscoNet \cite{DiscoNet} & 54.71 & 0.700 & 0.558 \\
\midrule
\rowcolor{gray!20} {\modelname} (ours) & \textbf{0.04} & 0.693 & 0.597 \\
\bottomrule
\end{tabular}
\end{adjustbox}
\end{table}

\begin{table}[htbp]
\caption{Object tracking results on the UAV3D dataset.}
\label{table: track}
\centering
\begin{adjustbox}{width=\columnwidth}
\begin{tabular}{c|c|c|c|c|c|c} 
\toprule
\multirow{2}{*}{\textbf{Method}} 
 & \textbf{AMOTA$\uparrow$} & \textbf{AMOTP$\downarrow$} & \textbf{MOTA$\uparrow$} & \textbf{MOTP$\downarrow$} & \textbf{TID$\downarrow$} & \textbf{LGD$\downarrow$} \\ \cline{2-7} 
\textbf   & (\%)  & (m) & (\%) & (m) & (s) & (s) \\ \hline 
\addlinespace[2.5pt]
No Collaboration     &0.644   &1.018     &0.593   &0.611	 &0.620    &1.280	\\
Early Collaboration  &\textbf{0.812}   &\textbf{0.672}     &\textbf{0.781}   &0.476	 &\textbf{0.300}    &\textbf{0.570} \\
\midrule
When2Com \cite{When2com}       &0.646   &1.012     &0.595   &0.618	&0.590    &1.200	\\ 
Who2Com \cite{Who2com}       &0.648   &1.012    &0.602   &0.623	&0.580    &1.200	\\ 
V2VNet \cite{V2Vnet}       &0.782   &0.803     &0.735   &0.587	&0.360    &0.710	\\
Where2comm \cite{Where2comm}       &0.793   &0.745     &0.751   &0.532	&0.340    &0.640	\\
DiscoNet \cite{DiscoNet}       &0.809   &0.703     &0.766   &0.516	&0.300    &0.590	\\  
\midrule

\rowcolor{gray!20} {\modelname} (ours) &0.744 &0.743 &0.678 &\textbf{0.471} &0.436 &0.867 \\
\midrule
\end{tabular}
\end{adjustbox}
\end{table}

\subsection{Experimental settings}
\textbf{Datasets and implementation details}.
Our experiments utilize two open-source multi-UAV collaborative datasets, UAV3D \cite{UAV3D} and Air-Co-Pred \cite{DHD}.
For a comprehensive evaluation, we conduct various downstream tasks, including 3D object detection and object tracking on the UAV3D dataset, and trajectory prediction on the Air-Co-Pred dataset.
The perception area is set to 204.8m $\times$ 204.8m in UAV3D and 100m $\times$ 100m in Air-Co-Pred.
To obtain the virtual modality Poture, we utilize three historical frames (i.e., $K=3$). 
All the models are trained on four NVIDIA GeForce RTX A6000 GPUs.

\par

\textbf{Evaluation metrics}.
For the 3D object detection task, we use mean Average Precision (mAP) and nuScenes Detection Score (NDS) as evaluation metrics.
We employ the object tracking metrics defined in UAV3D \cite{UAV3D}, including AMOTA, AMOTP, MOTA, MOTP, TID, and LGD.
In the trajectory prediction task, we use Intersection-over-Union (IoU) and Video Panoptic Quality (VPQ) for frame-level and video-level evaluation, respectively.

\begin{figure}[htbp] 
    \centering 
    \begin{subfigure}[b]{0.45\linewidth} 
        \centering
        \includegraphics[width=\linewidth]{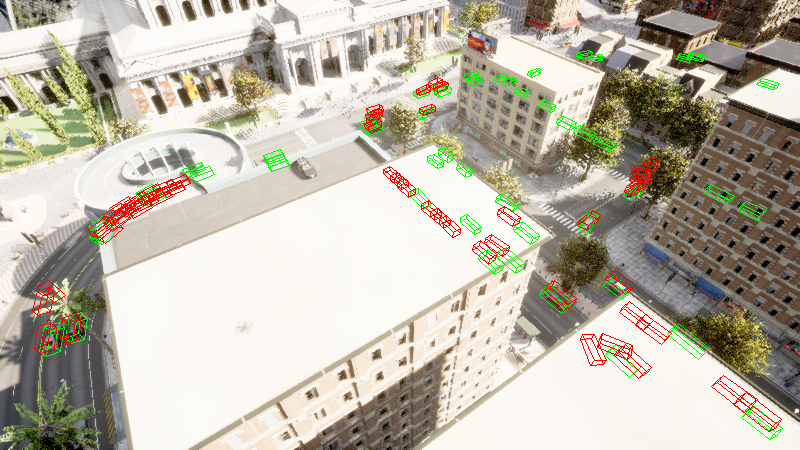} 
        \caption{No-fusion}
        \label{fig:sub1}
    \end{subfigure}
    \hfill 
    \begin{subfigure}[b]{0.45\linewidth}
        \centering
        \includegraphics[width=\linewidth]{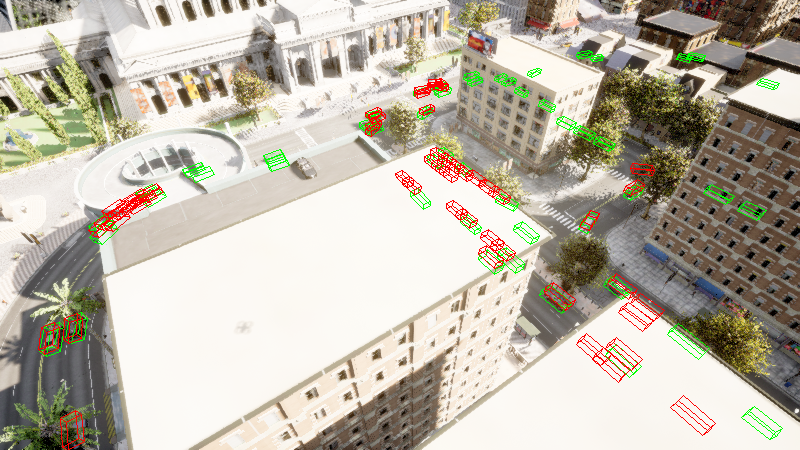}
        \caption{When2com}
        \label{fig:sub2}
    \end{subfigure}
    \begin{subfigure}[b]{0.45\linewidth}
        \centering
        \includegraphics[width=\linewidth]{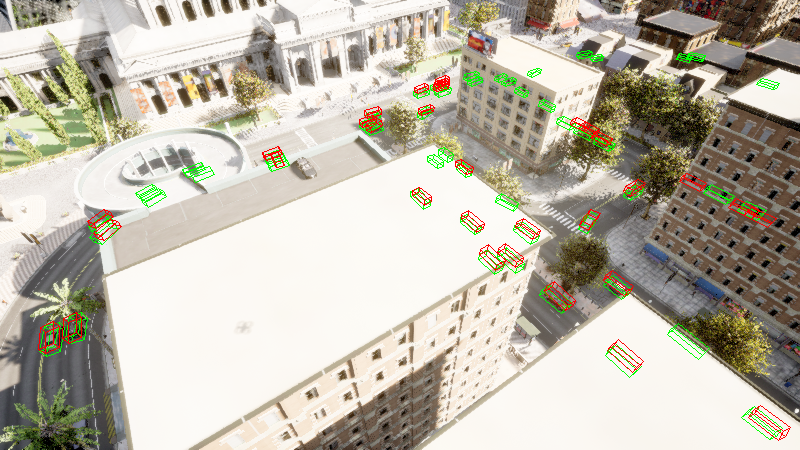}
        \caption{DiscoNet}
        \label{fig:sub3}
    \end{subfigure}
    \hfill
    \begin{subfigure}[b]{0.45\linewidth}
        \centering
        \includegraphics[width=\linewidth]{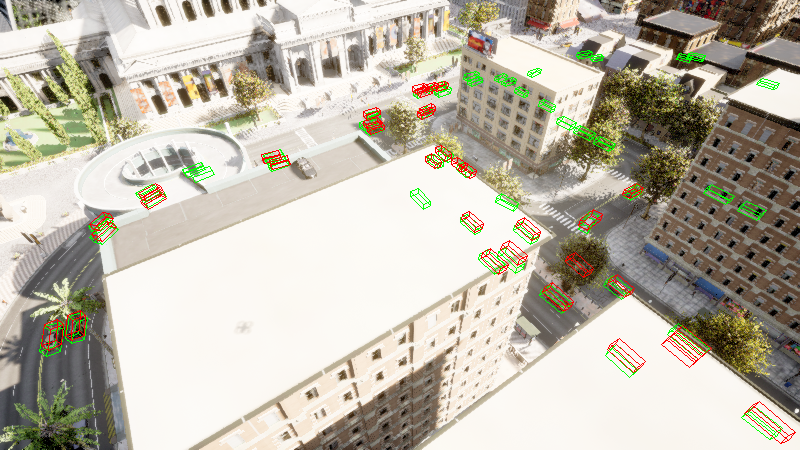}
        \caption{CERF (ours)}
        \label{fig:sub4}
    \end{subfigure}
    
    \caption{Visualization of 3D object detection results on the UAV3D dataset. Red and green boxes denote the predictions and ground truth, respectively.}
    \label{figure: visualization}
\end{figure}

\begin{table}[t!]
\renewcommand{\arraystretch}{1.1}
\caption{Trajectory prediction results on the Air-Co-Pred dataset. \textcolor[RGB]{180,180,180}{DHD} denotes the official results, and \text{*} denotes the results we reproduce.}
\centering
    \begin{adjustbox}{width=\linewidth}

    \begin{tabular}{c|c|c|c|c}
        
        \toprule
        \textbf{Method} & \textbf{Shared data}& \textbf{Retraining-free} & \textbf{IoU $\uparrow$} & \textbf{VPQ $\uparrow$} \\
        \midrule
        No Collaboration & / & / & 0.326 & 0.278 \\ 
        Early Collaboration & RGB images & \checkmark & 0.545 &  0.459 \\ 
        Late Collaboration & perception outputs & \checkmark & 0.514 & 0.436 \\
        \cline{1-5}
        \addlinespace[2pt]
        V2X-ViT\cite{V2X-ViT} & feature maps & \ding{55} &  0.533 &  0.457 \\
        V2VNet\cite{V2Vnet} & feature maps & \ding{55} &  0.538 &  0.469 \\ 
        Who2com\cite{Who2com} & feature maps & \ding{55} & 0.443 & 0.370 \\
        When2com\cite{When2com} & feature maps & \ding{55} & 0.458 & 0.404 \\
        Where2comm\cite{Where2comm} & feature maps & \ding{55} & 0.514 & 0.442 \\
        UMC\cite{UMC} & feature maps & \ding{55}  & 0.523 & 0.443 \\
        UMC (w/o GRU) & feature maps & \ding{55}  & 0.532 & 0.459 \\
        \textcolor[RGB]{180,180,180}{DHD\cite{DHD}} & \textcolor[RGB]{180,180,180}{feature maps} & \textcolor[RGB]{180,180,180}{\ding{55}} & \textcolor[RGB]{180,180,180}{0.540} & \textcolor[RGB]{180,180,180}{0.462} \\
        {DHD$^\ast$} & feature maps & \ding{55} & \textbf{0.532}  & \textbf{0.460}  \\ 
        \cline{1-5}
        \rowcolor{gray!20} {\modelname} (ours) & perception outputs & \checkmark & 0.530 & 0.458 \\ 
        \bottomrule
    \end{tabular}
    \end{adjustbox}
\label{table: DHD}
\vspace{-8pt}
\end{table}

\subsection{Quantitative evaluation}
\textbf{Performance comparison.}
Table \ref{table: overall_performace} presents the performance comparison of 3D object detection on the UAV3D dataset. Experimental results show that our {\modelname} outperforms most intermediate-collaboration methods while requiring only about 1/1000 of the communication bandwidth (comparable to late-fusion). Table \ref{table: track} shows the object tracking performance of our framework and previous methods. It can be observed that, even by transmitting and fusing only compact perception results, our {\modelname} still achieves acceptable tracking accuracy.
\par
To evaluate the generalization ability of our method across various datasets and tasks, we show the trajectory prediction results on the Air-Co-Pred dataset in Table \ref{table: DHD}. We see that {\modelname} achieves 0.530 in IoU and 0.458 in VPQ, which are on par with those of existing state-of-the-art methods.
The effectiveness of the CERF framework can be attributed to its ability to extract sufficient complementary cues from perception results in a learnable fashion.
\par

\textbf{Open heterogeneity evaluation.}
Table \ref{table: hetero} illustrates the performance of our {\modelname} under open heterogeneous settings. Only Agent 2 participates in collaborative training with the ego agent, while Agent 1 and Agent 3 appear exclusively at the inference stage. The results indicate that {\modelname} can incorporate informative data from unknown agents without retraining cost, independent of input resolution, backbone architecture, or detection threshold. The collaborative performance is primarily determined by the capability of individual perception. For example, since Agent 3 achieves the highest single-agent perception accuracy, the resulting collaborative network also delivers the strongest perceptual performance.

\begin{table}[htbp]
\caption{Collaborative performance of {\modelname} under open heterogeneous settings. \text{*} denotes participation in the training phase, and $^\circledast$ denotes occurrence only in the inference phase.}
\label{table: hetero}

\centering
\begin{adjustbox}{width=\columnwidth}
\begin{tabular}{cccccc}
\toprule
\textbf{Setting} & \textbf{Encoder} & \textbf{Input size} & \textbf{Threshold} & \textbf{mAP $\uparrow$} & \textbf{NDS $\uparrow$} \\

\hline
\addlinespace[2.5pt]
Ego agent & BEVFusion \cite{BEVFusion} & $800\times450$ & 0.25 & 0.544 & 0.481 \\ 
Agent 1 & BEVFusion & $704\times256$ & 0.25 & 0.487 & 0.458 \\
Agent 2 & PETR \cite{PETR} & $800\times450$ & 0.2 & 0.581 & 0.516 \\
Agent 3 & DETR3D \cite{DETR3D} & $800\times450$ & 0.3 & 0.618 & 0.547 \\
\midrule
Ego + Agent 1$^\circledast$ & / & / & / & \cellcolor{gray!20}0.613 & \cellcolor{gray!20}0.523 \\

Ego + Agent 2$^\ast$ & / & / & / & \cellcolor{gray!20}0.660 & \cellcolor{gray!20}0.572 \\
Ego + Agent 3$^\circledast$ & / & / & / & \cellcolor{gray!20}\textbf{0.693} & \cellcolor{gray!20}\textbf{0.597} \\
\bottomrule
\end{tabular}
\end{adjustbox}
\end{table}

\begin{table}[htbp]
\caption{Effects of different object attributes in Poture.}
\label{table: ablation}
\centering
\begin{adjustbox}{width=\columnwidth} 
\begin{tabular}{c c c c| c c}
\toprule
 \textbf{Size} & \textbf{Heading angle} & \textbf{Confidence score} & \textbf{Location} & \textbf{mAP$\uparrow$} & \textbf{NDS$\uparrow$} \\
\midrule
\addlinespace[2.5pt]
 \ding{55} & \ding{55} & \ding{55} &  \ding{55} & 0.544 & 0.450 \\ 
 \checkmark & \ding{55} & \ding{55} & \ding{55} & 0.628 & 0.551 \\ 
 \checkmark & \checkmark & \ding{55} & \ding{55} & 0.644 & 0.565 \\ 
 \rowcolor{gray!20} \checkmark & \checkmark & \checkmark & \ding{55} & \textbf{0.693} & \textbf{0.597} \\ 
 \checkmark & \checkmark  & \checkmark & \checkmark & 0.650 & 0.568 \\ 
\midrule
\end{tabular}
\end{adjustbox}
\end{table}
\vspace{-8pt}

\subsection{Qualitative evaluation}
Fig. \ref{figure: visualization} presents the visualization of 3D object detection results on the UAV3D dataset.
Our {\modelname} achieves bounding box accuracy comparable to mainstream intermediate-collaboration methods, and even surpasses them in certain scenarios.
This demonstrates that {\modelname} can effectively aggregate complementary information from other agents' perception outputs, enhancing practicality while maintaining performance.

\subsection{Ablation studies}
Table \ref{table: ablation} shows the importance of different attributes in the proposed virtual modality Poture. Box size, heading angle, and confidence score help improve perception, whereas incorporating the center location in Poture leads to performance degradation. The reason is that the corresponding positions in Poture are represented by discretized BEV grids, making world-coordinate positions unnecessary.
\section{Conclusion}
This paper introduces {\modelname}, a novel communication-efficient
and retraining-free framework for multi-UAV collaboration.
{\modelname} generates the virtual modality Poture to enhance the local BEV features, sharing only compact perception results among connected agents.
To mitigate transmission delays, we use a Kalman-filter based tracker and motion forecasting to derive current predictions from the historical messages.
Experimental results show that {\modelname} achieves competitive performance with minimal bandwidth, and seamlessly integrates unknown agents into collaboration.

\newpage
\section{ACKNOWLEDGMENT}
This work was supported by the National Science and Technology Major Project under Grant 2022ZD0116409.

\bibliographystyle{IEEEbib}
\bibliography{references}

@article{UAV3D,
  title={UAV3D: A Large-scale 3D Perception Benchmark for Unmanned Aerial Vehicles},
  author={Ye, Hui and Sunderraman, Rajshekhar and others},
  journal={arXiv preprint arXiv:2410.11125},
  year={2024}
}

@inproceedings{V2X-ViT,
  title={V2x-vit: Vehicle-to-everything cooperative perception with vision transformer},
  author={Xu, Runsheng and Xiang, Hao and others},
  booktitle={ECCV},
  pages={107--124},
  year={2022},
}

@article{DHD,
  title={Drones Help Drones: A Collaborative Framework for Multi-Drone Object Trajectory Prediction and Beyond},
  author={Wang, Zhechao and Cheng, Peirui and others},
  journal={arXiv preprint arXiv:2405.14674},
  year={2024}
}

@inproceedings{BEVFusion,
  title={Bevfusion: Multi-task multi-sensor fusion with unified bird's-eye view representation},
  author={Liu, Zhijian and Tang, Haotian and others},
  booktitle={ICRA},
  pages={2774--2781},
  year={2023},
}

@inproceedings{When2com,
  title={When2com: Multi-agent perception via communication graph grouping},
  author={Liu, Yen-Cheng and Tian, Junjiao and others},
  booktitle={CVPR},
  pages={4106--4115},
  year={2020}
}

@inproceedings{Who2com,
  title={Who2com: Collaborative perception via learnable handshake communication},
  author={Liu, Yen-Cheng and Tian, Junjiao and others},
  booktitle={ICRA},
  pages={6876--6883},
  year={2020},
}

@inproceedings{V2Vnet,
  title={V2vnet: Vehicle-to-vehicle communication for joint perception and prediction},
  author={Wang, Tsun-Hsuan and Manivasagam, Sivabalan and others},
  booktitle={ECCV},
  pages={605--621},
  year={2020},
}

@article{DiscoNet,
  title={Learning distilled collaboration graph for multi-agent perception},
  author={Li, Yiming and Ren, Shunli and others},
  journal={NIPS},
  pages={29541--29552},
  year={2021}
}

@inproceedings{MoDAR,
  title={Modar: Using motion forecasting for 3d object detection in point cloud sequences},
  author={Li, Yingwei and Qi, Charles R and others},
  booktitle={CVPR},
  pages={9329--9339},
  year={2023}
}

@article{HEAL,
  title={An extensible framework for open heterogeneous collaborative perception},
  author={Lu, Yifan and Hu, Yue and others},
  journal={arXiv preprint arXiv:2401.13964},
  year={2024}
}

@article{Where2comm,
  title={Where2comm: Communication-efficient collaborative perception via spatial confidence maps},
  author={Hu, Yue and Fang, Shaoheng and others},
  journal={NIPS},
  pages={4874--4886},
  year={2022}
}

@inproceedings{UMC,
  title={Umc: A unified bandwidth-efficient and multi-resolution based collaborative perception framework},
  author={Wang, Tianhang and Chen, Guang and others},
  booktitle={ICCV},
  pages={8187--8196},
  year={2023}
}

@article{Griffin,
  title={Griffin: Aerial-Ground Cooperative Detection and Tracking Dataset and Benchmark},
  author={Wang, Jiahao and Cao, Xiangyu and others},
  journal={arXiv preprint arXiv:2503.06983},
  year={2025}
}

@article{QuantV2X,
  title={QuantV2X: A Fully Quantized Multi-Agent System for Cooperative Perception},
  author={Zhao, Seth Z and Zhang, Huizhi and others},
  journal={arXiv preprint arXiv:2509.03704},
  year={2025}
}

@article{yang2025discretization,
  title={Is Discretization Fusion All You Need for Collaborative Perception?},
  author={Yang, Kang and Bu, Tianci and others},
  journal={arXiv preprint arXiv:2503.13946},
  year={2025}
}

@inproceedings{IFTR,
  title={IFTR: An instance-level fusion transformer for visual collaborative perception},
  author={Wang, Shaohong and Bin, Lu and others},
  booktitle={ECCV},
  pages={124--141},
  year={2024},
}

@article{STAMP,
  title={Stamp: Scalable task and model-agnostic collaborative perception},
  author={Gao, Xiangbo and Xu, Runsheng and others},
  journal={arXiv preprint arXiv:2501.18616},
  year={2025}
}

@inproceedings{PETR,
  title={Petr: Position embedding transformation for multi-view 3d object detection},
  author={Liu, Yingfei and Wang, Tiancai and others},
  booktitle={ECCV},
  pages={531--548},
  year={2022},
}

@inproceedings{DETR3D,
  title={Detr3d: 3d object detection from multi-view images via 3d-to-2d queries},
  author={Wang, Yue and Guizilini, Vitor Campagnolo and others},
  booktitle={CoRL},
  pages={180--191},
  year={2022},
}

@article{KF_tracker,
  title={A baseline for 3d multi-object tracking},
  author={Weng, Xinshuo and Kitani, Kris},
  journal={arXiv preprint arXiv:1907.03961},
  pages={6},
  year={2019}
}

@inproceedings{TransIFF,
  title={Transiff: An instance-level feature fusion framework for vehicle-infrastructure cooperative 3d detection with transformers},
  author={Chen, Ziming and Shi, Yifeng and others},
  booktitle={ICCV},
  pages={18205--18214},
  year={2023}
}

@article{AGC-Drive,
  title={AGC-Drive: A Large-Scale Dataset for Real-World Aerial-Ground Collaboration in Driving Scenarios},
  author={Hou, Yunhao and Zou, Bochao and others},
  journal={arXiv preprint arXiv:2506.16371},
  year={2025}
}

@inproceedings{U2UData,
  title={U2udata: A large-scale cooperative perception dataset for swarm uavs autonomous flight},
  author={Feng, Tongtong and Wang, Xin and others},
  booktitle={Proceedings of the 32nd ACM International Conference on Multimedia},
  pages={7600--7608},
  year={2024}
}

@inproceedings{Hetecooper,
  title={Hetecooper: Feature collaboration graph for heterogeneous collaborative perception},
  author={Shao, Congzhang and Luo, Guiyang and others},
  booktitle={ECCV},
  pages={162--178},
  year={2024},
}

@inproceedings{OPV2V,
  title={Opv2v: An open benchmark dataset and fusion pipeline for perception with vehicle-to-vehicle communication},
  author={Xu, Runsheng and Xiang, Hao and others},
  booktitle={ICRA},
  pages={2583--2589},
  year={2022},
}

\end{document}